\documentclass[conference, letterpaper]{IEEEtran}
\IEEEoverridecommandlockouts
\usepackage[noend,ruled,linesnumbered]{algorithm2e}
\usepackage{cite}
\usepackage{amsmath,amssymb,amsfonts}
\usepackage{algorithmic}
\usepackage{graphicx}
\usepackage{textcomp}
\usepackage{xcolor}
\usepackage{caption}
\usepackage{subcaption}
\usepackage{hyperref}
\usepackage{multirow}
\usepackage[letterpaper, top=1.92cm, bottom=3.2cm, left=1.92cm, right=1.92cm]{geometry}

\newtheorem{theorem}{Theorem}
\newtheorem{definition}{Definition}

\newtheorem{lemma}{Lemma}

\newtheorem{assumption}{Assumption}

\usepackage{threeparttable} 
\usepackage{booktabs}

\def\BibTeX{{\rm B\kern-.05em{\sc i\kern-.025em b}\kern-.08em
    T\kern-.1667em\lower.7ex\hbox{E}\kern-.125emX}}
\begin{document}

\title{ALI-DPFL: Differentially Private Federated Learning with Adaptive Local Iterations
}


\author{  
\IEEEauthorblockN{Xinpeng Ling*, Jie Fu*, \thanks{*: These authors contributed equally.}Kuncan Wang, Haitao Liu, Zhili Chen\dag\thanks{\dag: Corresponding author.}}  
\IEEEauthorblockA{\textit{Shanghai Key Laboratory of Trustworthy Computing, East China Normal University, Shanghai, China} \\  
$\{$xpling, jie.fu, kuncan.wang, htliu$\}$@stu.ecnu.edu.cn, zhlchen@sei.ecnu.edu.cn
}
}

\maketitle

\begin{abstract}
Federated Learning (FL) is a distributed machine learning technique that allows model training among multiple devices or organizations by sharing training parameters instead of raw data. However, adversaries can still infer individual information through inference attacks (e.g. differential attacks) on these training parameters. As a result, Differential Privacy (DP) has been widely used in FL to prevent such attacks.

We consider differentially private federated learning in a resource-constrained scenario, where both privacy budget and communication rounds are constrained. 
By theoretically analyzing the convergence, we can find the optimal number of local Differential Privacy Stochastic Gradient Descent (DPSGD) iterations for clients between any two sequential global updates. Based on this, we design an algorithm of \underline{D}ifferentially \underline{P}rivate \underline{F}ederated \underline{L}earning with \underline{A}daptive \underline{L}ocal \underline{I}terations (ALI-DPFL). We experiment our algorithm on the MNIST, FashionMNIST and Cifar10 datasets, and demonstrate significantly better performances than previous work in the resource-constraint scenario. Code is available at \href{https://github.com/cheng-t/ALI-DPFL}{https://github.com/cheng-t/ALI-DPFL}.
\end{abstract}


\section{Introduction}
\addtolength{\topmargin}{0.03in}

In federated learning, each client normally downloads the global model from the center sever, performs local iterations, and uploads the resulted training parameters back to the center sever. The center sever then updates the global model accordingly. The above steps repeat until the global model converges \cite{mcmahan2017communication}. In this way, the global model reaches convergence by communicating only training parameters instead of raw data. Nevertheless, some studies have shown that federated learning still carries privacy risks. Training parameters, such as gradient values, can be used to recover a portion of the original data \cite{zhu2019deep} or infer whether specific content originates from certain data contributors \cite{song2017machine}. Melis et al. \cite{LucaMelis2022ExploitingUF} also demonstrated that participants’ training data could be leaked by shared models. Therefore, additional measures need to be taken to protect data privacy in federated learning.

Many works\cite{ AnttiKoskela2018LearningRA,LeTrieuPhong2017PrivacyPreservingDL,jayaraman2019evaluating,LiyaoXiang2019DifferentiallyPrivateDL} have demonstrated that the technique of differential privacy (DP) 
could protect machine learning models from unintentional information leakage. 
In differentially private federated learning (DPFL), each client executes a certain number of local stochastic gradient descent with differential privacy (DPSGD) iterations before performing a global aggregation. 
For each global aggregation, DPFL consumes privacy budget according to the number of local iterations and communicates one round of training parameters, and thus DPFL should be constrained by both the privacy budget and communication rounds.
However, current DPFL schemes typically optimize the global model only with the constraint of a limited privacy budget, while overlooking the constraint of communication rounds.

\begin{figure}[bht]  
    \centering  
    \begin{minipage}[b]{0.14\textwidth}  
        \centering  
        \includegraphics[width=\textwidth]{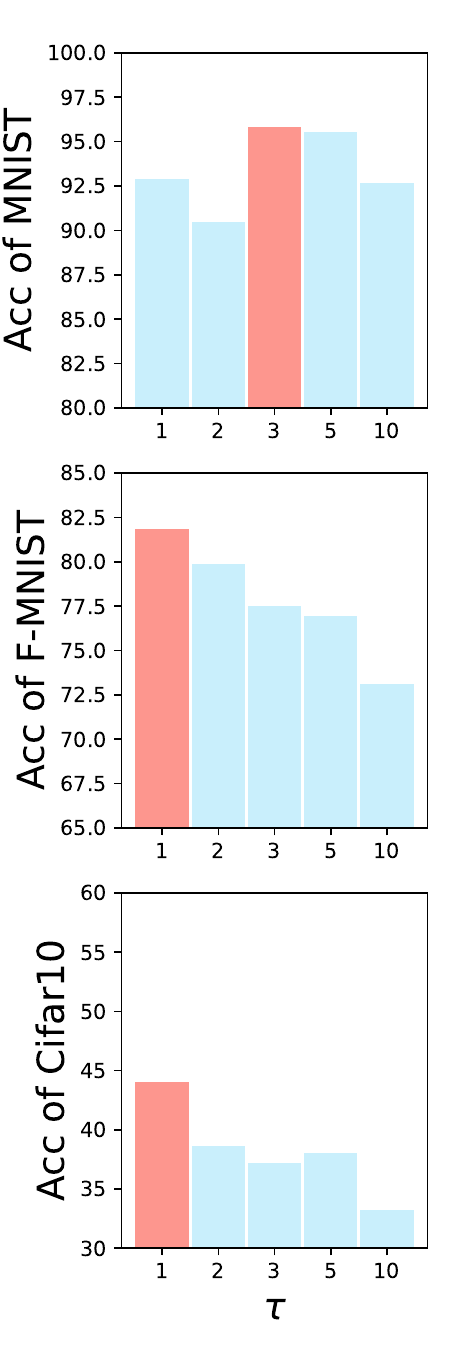}  
        \subcaption{Iters=150}  
        \label{fig:00_1_Rs2Rc}  
    \end{minipage}  
    \hfill  
    \begin{minipage}[b]{0.15\textwidth}  
        \centering  
        \includegraphics[width=\textwidth]{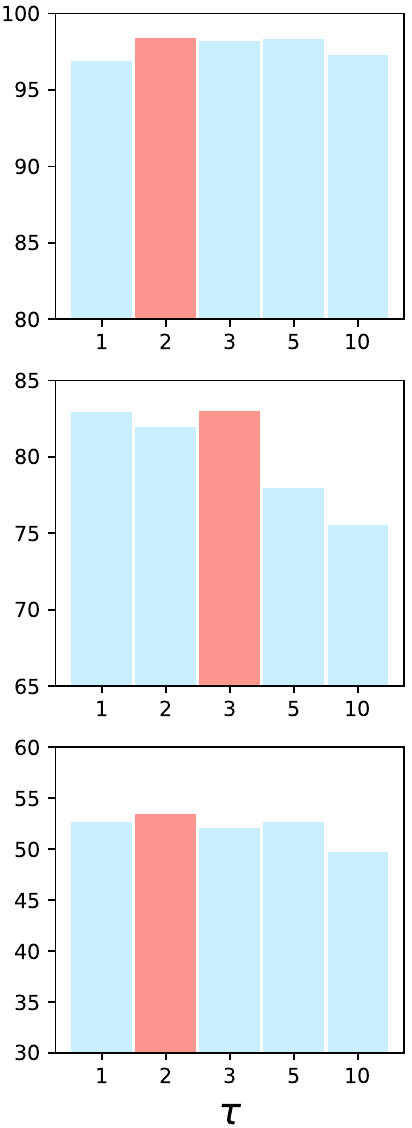}  
        \subcaption{Iters=300}  
        \label{fig:00_3_Rs12Rc}  
    \end{minipage}  
    \hfill  
    \begin{minipage}[b]{0.15\textwidth}  
        \centering  
        \includegraphics[width=\textwidth]{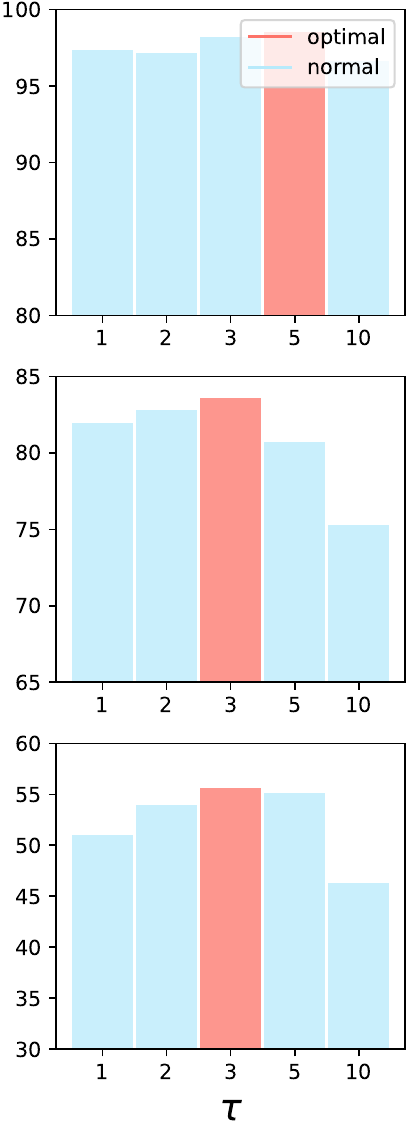}  
        \subcaption{Iters=600}  
        \label{fig:00_4_Rs15Rc}  
    \end{minipage}  
  \caption{The test accuracy($\%$) of fixed $\tau$ in different datasets and iterations.} 
    \label{fig:fixed_tau_performace}  
\end{figure}


\begin{figure*}[thb]  
  \centering  
  \includegraphics[width=1.0\textwidth]{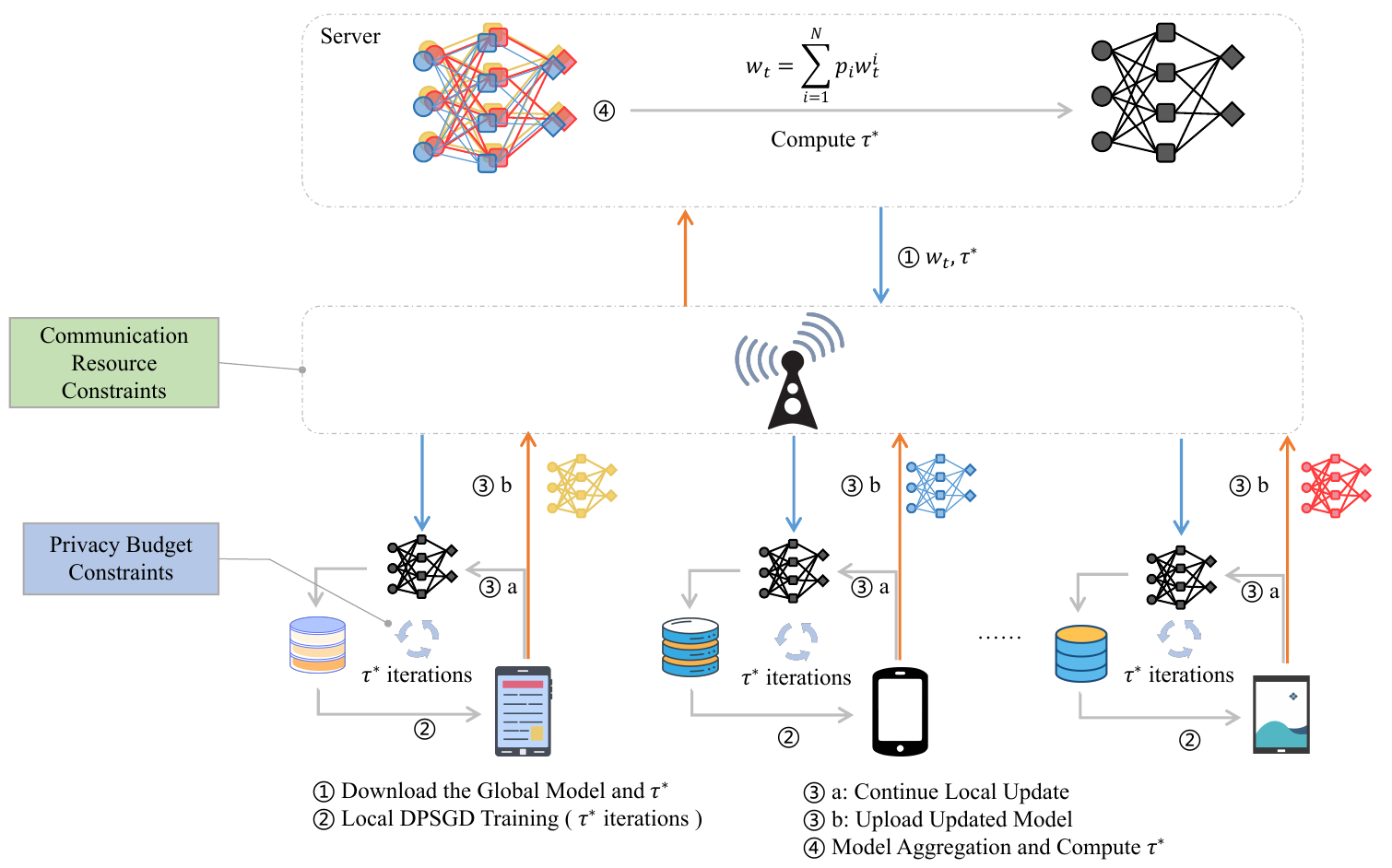}  
  \caption{The Framework of Differentially Private Federated Learning with Adaptive Local Iterations}
  \label{fig:FL}  
\end{figure*}

Existing DPFL schemes usually choose empirically a fixed number of local iterations $\tau$ for each global update, usually, $\tau = 1$.  
They treat communication rounds as unlimited, but in practice, the number of communication rounds are often constrained by limited resources such as time and bandwidth. As shown in Fig. \ref{fig:fixed_tau_performace}, the optimal fixed $\tau$ value corresponding to the highest accuracy is not a fixed value across different datasets and iterations. It is also indicated that a fixed $\tau$ may fail to achieve satisfactory convergence performances \cite{LiT2020FedProx,WangJ2020FedNova,WangS2019ResourceConstrained}. Therefore, for the first time, in this paper we focus on finding a DPFL scheme with adaptive local iterations to achieve good performances when both privacy budget and communication rounds are constrained.

Through the convergence analysis, we derive a convergence bound related to the number $\tau$ of local iterations. Then, we propose a differentially private federated learning scheme with adaptive local iterations (ALI-DPFL), as shown in Fig. \ref{fig:FL}, which finds the optimal value ($\tau^*$) in each communication round that minimizes the convergence bound. This value is then used as the number of local iterations in the next round, enabling the algorithm to converge faster under the constraints of privacy budget and communication rounds. The main contributions of this paper are as follows:


\begin{enumerate}  
    \item We analyze DPFL convergence under constrained privacy budgets and communication rounds, presenting a novel convergence bound related to the number of local iterations.  
    \item Leveraging this bound, we introduce the ALI-DPFL algorithm, employing adaptive local iterations for high accuracy and fast convergence in resource-constrained settings.
    \item Our extensive experiments showcase our approach's superiority over three leading DPFL algorithms across IID and Non-IID data distributions..
\end{enumerate}  


The rest of the paper is organized as follows. Section \ref{sec:related_work} reviews prior research. Section \ref{sec:preliminaries} introduces the preliminary knowledge. Section \ref{sec:ali_dpfl_algorithm} proposes the ALI-DPFL algorithm through analyzing convergence bound of the DPFL algorithm and proves our algorithm satisfy differential privacy. The experimental results are presented in Section \ref{sec:experiment}, followed the conclusion in Section \ref{sec:conclusion}.

\section{Related Work}\label{sec:related_work}
\addtolength{\rightmargin}{0.04in}

\subsection{DPFL}
In previous studies, two different types of differential privacy have been incorporated into federated learning to protect privacy: client-level differential privacy (CL-DP) and sample-level differential privacy (SL-DP).

\textbf{CL-DP.} In the CL-DP setting, the model's performance remains stable despite the addition or removal of one client during training \cite{HBrendanMcMahan2017LearningDP}. Enayat et al. \cite{Enayat2023PrivateFLwithAutotunedCompression} utilized CL-DP to dynamically adjust the compression rate based on training loss, thereby reducing communication rounds in federated learning. Zhou et al. \cite{zhou2023exploring} explored the impact of local iterations when employing the Laplace mechanism for perturbation.


\textbf{SL-DP.} In the SL-DP setting, each client possesses a set of data samples. Importantly, the addition or removal of a data sample during training has minimal impact on the model’s performance \cite{MartnAbadi2016DeepLW,XiWu2016BoltonDP}. Fu et al. \cite{FuJ2022AdapDPFL} introduced adaptive gradient clipping, tailoring it for different clients and communication rounds based on observed changes in loss values. Raouf et al. \cite{raouf2021ConstrainedDPFLforlowBandwidth} leveraged SL-DP to enhance model accuracy by updating only a fixed subset of model weights while leaving others unchanged, thereby reducing bandwidth consumption for both upstream and downstream communication. Kang Wei et al. \cite{KangWei2021UserLevelPF} proposed the adaptive privacy budget allocation. Zhou et al. \cite{zhou2021optimizing} provided the optimal value for the total local iterations through convergence analysis and analyzed the impact of the total number of iterations on convergence performance and privacy protection. Our work focuses on the SL-DP setting, which is the most popular in the real world.

\subsection{Adaptive local iterations in FL}
There are currently numerous adaptive federated learning algorithms aimed at improving model performance in heterogeneous data settings. For example adaptive model aggregation \cite{Sunwoo2023LayerWise, jayaram2022adaptive}, adaptive optimizers \cite{reddi2020FedOPT}, adaptive Network Layering~\cite{karimi2021layer} and adaptive client sampling~\cite{Boxin2023AddressingBudget}, etc. Wang et al\cite{WangS2019ResourceConstrained} improved the federated learning algorithm by considering scenarios where clients and the server consume the same type of resources. They achieved faster convergence of the global model under resource constraints by dynamically adjusting local iterations.

Some studies like \cite{WangS2019ResourceConstrained} addressed optimal iteration rounds in FL under limited resources but overlooked differential privacy. Others, such as \cite{zhou2023exploring}\cite{zhou2021optimizing}, calculated optimal iteration rounds in DPFL but neglected communication round constraints. This paper, to our knowledge, represents the first attempt to utilize SL-DP convergence analysis to dynamically determine the number of local iterations, achieving rapid convergence.

\section{Preliminaries}\label{sec:preliminaries}
\addtolength{\topmargin}{0.03in}
In the beginning, we list some notations used in this paper in Table \ref{table:notations}. Symbols that have not been mentioned before will be defined later in the following sections.

\begin{table}[htb]
\centering
\caption{Summary of main notations}
\label{table:notations}
\begin{tabular}{p{0.05\textwidth} p{0.38\textwidth}}
\hline
$F(\mathbf{w}$) & Global loss function  \hfill\\
$F_i(\mathbf{w}$) & Local loss function for client $i$  \hfill\\
$t$ & Iteration index  \hfill\\
$k$ & Global aggregation index  \hfill\\
$\mathbf{w}^i_t$ & Local model parameter at node $i$ at iteration $t$  \hfill\\
$\mathbf{w}_t$ & Global model parameter at iteration $t$  \hfill\\
$\mathbf{w}^*$ & True optimal model parameter that minimizes $F(\mathbf{\mathbf{w}})$ \hfill\\
$\eta$ & Gradient descent step size \hfill\\
$\tau$ & Number of local update iterations between two global aggregations\hfill\\
$\tau^*$ & Dynamic optimal $\tau$ in a single round\hfill\\
$T$ & Total number of local iterations\hfill\\
$K$ & Total number of global aggregation, equal to $T/\tau$\hfill\\
$p_i$ & weight of client $i$ \hfill\\
$R_s$ & Maximum communication rounds \hfill\\
$R_c$ & Maximum total local iterations \hfill\\
$L$ & Lipschitz parameter\hfill\\
$\mu$ & Strongly convex parameter\hfill\\
$\sigma$ & Standard deviation of Gaussian distribution\hfill \\
$d$ & The dimension of model\hfill \\
$C$ & Clipping bound\hfill \\
$q$ & Sampling rate of batch\hfill \\
$||\cdot||$ & $L_2$-norm \hfill \\
\hline
\end{tabular}
\end{table}
\subsection{Differential privacy}
Differential privacy is a rigorous mathematical framework that formally defines the privacy loss of data analysis algorithms. Informally, it requires that any changes to a single data point in the training dataset can not cause statistically significant changes in the algorithm's output.

\begin{definition}
	(Differential privacy\cite{dwork2014algorithmic}). $(\epsilon, \delta)$-Differential privacy is achieved by a randomized mechanism $\mathcal{M}: \mathcal{X} \rightarrow \mathcal{R}$ if, for any two neighboring databases $\mathcal{D}_i, \mathcal{D}_i^{\prime} \in \mathcal{X}$ that differ in only a single entry, and $\forall$ $\mathcal{S} \subseteq \text{Range}(\mathcal{R})$,
\end{definition}
\begin{equation}\label{def:DP}
\operatorname{Pr}\left[\mathcal{M}\left(\mathcal{D}_i\right) \in \mathcal{S}\right] \leq e^\epsilon \operatorname{Pr}\left[\mathcal{M}\left(\mathcal{D}_i^{\prime}\right) \in \mathcal{S}\right]+\delta
\end{equation}





\subsection{Differentially private stochastic gradient descent}
Differentially Private Stochastic Gradient Descent (DPSGD) stands as a widely embraced deep learning algorithm offering privacy guarantees\cite{MartnAbadi2016DeepLW,MingyiHong2020UnderstandingGC,ZhiqiBu2022Automatic,XiaodongYang2022Normalized,TianyuXia2023DPLwithPAC}. Specifically, at each iteration $t$, a batch of data $\mathcal{B}_t$ is sampled from the dataset using a sampling rate $q$. Subsequently, for each batch, $s$ is a sample thereof. Following the clipping of per-sample gradients, Gaussian noise with a mean of 0 is incorporated into the gradients in the subsequent step:

\begin{align}
\bar {g}_t(s) &= \textbf{clip}\left(\nabla F(\mathbf{w}_t,s)\right) \nonumber\\
&= \nabla F(\mathbf{w}_t,s) / \max \left(1, \frac{\|\nabla F(\mathbf{w}_t,s)\|}{C}\right)
\end{align}

\begin{align}
\tilde g_t = \frac{1}{|\mathcal{B}_t|}\left( \sum_{s \in \mathcal{B}_t} \bar {g}_t(s) + \mathcal{N}(0,\sigma^2 C^2 \mathbf{I}_d) \right)
\end{align}

Here, $\sigma$ represents the noise multiplier, while $C$ denotes the clipping bound. Subsequently, gradient descent is executed based on the batch-averaged gradients. The initial models being randomly generated and independent of the sample data, combined with the batch-averaged gradients adhering to differential privacy, ensures that the resulting models also adhere to differential privacy, attributable to the post-processing property\cite{jiefu2023DPSUR}.

\subsection{Federated Learning with Differential Privacy}
\begin{algorithm}
\caption{Federated Learning with Differential Privacy}\label{alg:DPFL}
\KwIn{loss function $F(\mathbf{w})$. Parameters: learning rate $\eta$, noise multiplier $\sigma$, local iterations $\tau$, clipping bound $C$, sampling rate $q$}
\KwOut{the final trained model $\mathbf{w}(K)$}
Initialize $k=0$, $\mathbf{w}_0=\text{Initial()}$\;
\While{$k<K$}
{
\For{$i\in [1,2,\cdots,N]$ parallel}{
\For{$j \in [1,2,\cdots,\tau]$}
{
$t=(k-1)\tau+j$\;
Sample randomly a batch $\mathcal{B}^i_t$ with sampling rate $q$\;
\For{$s\in \mathcal{B}^i_t$ }
{
$g^i_t(s) \leftarrow \nabla F_i(\mathbf{w}^i_t,s)$\; 
$\Bar{g}^i_t(s) \leftarrow g^i_t(s) / \max (1,\frac{\|g^i_t(s)\|}{C})$\; \label{alg1:line9}
}
$\widetilde{g}^i_t\leftarrow \sum\limits_{s \in {\mathcal{B}^i_t}}
\Bar{g}^i_t(s)+\mathcal{N}(0,\sigma^2 {C}^2 \mathbf{I}_d)$\;\label{alg1:line10}

$\mathbf{w}^i_{t+1}=\mathbf{w}^i_t-\eta \cdot \frac{\widetilde{g}^i_t}{|\mathcal{B}^i_t|}$\;
}
}
$\text{Receive } \mathbf{w}^i_{t+1}$\;
$k=k+1$\;
$\mathbf{w}(k)=\sum_{i=1}^N p_i \mathbf{w}^i_{t+1}$\;
$\text{Broadcast } \mathbf{w}(k) \text{ to each client}$ \;
}
\Return $\mathbf{w}(K)$
\end{algorithm}

In a typical federated learning system, a server and $N$ clients form the fundamental components\cite{mcmahan2017communication}\cite{mcmahan2016federated}. Each client, denoted as $\mathcal{C}_i$, maintains a local database $\mathcal{D}_i$, where $i \in \{1,2,\dots,N\}$. The quantity $|\mathcal{D}_i|$ signifies the size of the dataset at client $i$, and the total dataset size is denoted as $|\mathcal{D}|=\sum_{i=1}^{N} |\mathcal{D}_i|$. Let $p_i\triangleq\frac{|\mathcal{D}_i|}{|\mathcal{D}|}$ represent the weight of client $i$. The server's objective is to learn a model using data distributed across the $N$ clients. Each active client engaged in local training endeavors to discover a vector $\mathbf{w}$ that represents the neural network model minimizing a specific loss function. Subsequently, the server aggregates the weights from the $N$ clients, a process that can be expressed as:

\begin{equation}\label{eq:central_avg}
  \mathbf{w}(k)= \sum_{i=1}^N p_i \mathbf{w}^i_t
\end{equation}

Here, $\mathbf{w}^i_t$ signifies the parameter vector acquired from client $i$ at iteration $t$, while $\mathbf{w}(k)$ denotes the server-aggregated parameter vector at communication round $k$. The objective is to minimize $F(\mathbf{w})$, in other words, to seek:

\begin{equation}\label{eq:w_star}
\mathbf{w}^* \triangleq \arg \min F(\mathbf{w})
\end{equation}

The function $F(\cdot)$ is defined as follows:

\begin{equation}\label{eq:F_w}
F(\mathbf{w}) \triangleq \sum_{i=1}^N p_i F_i(\mathbf{w})
\end{equation}

Here, $F_i(\cdot)$ represents the local loss function of client $i$.

As depicted in Algorithm \ref{alg:DPFL}, the differentiating factor between the DPFL algorithm and traditional federated learning algorithms resides in the utilization of DPSGD instead of SGD. Specifically, at line \ref{alg1:line9}, the DPFL algorithm executes per-sample clipping, and at line \ref{alg1:line10}, it introduces noise to the clipped gradients.

\section{ALI-DPFL Algorithm}\label{sec:ali_dpfl_algorithm}
\addtolength{\topmargin}{0.03in}
In this section, we introduce our ALI-DPFL algorithm, which is based on the insights gleaned from the convergence analysis. In Section \ref{subsec:ca}, we conducted a convergence analysis of the traditional DPFL algorithm to investigate the relationship between the convergence upper bound and $\tau$, resulting in Equation (\ref{eq:ca}). Leveraging the outcomes of the convergence analysis, we derived Lemma \ref{lem:tau_0} and Lemma \ref{lem:tau_star}. Subsequently, in Section \ref{subsec:overview}, we outline the workflow of ALI-DPFL. Furthermore, in Section \ref{subsec:algo_alidpfl}, we constructed Algorithm \ref{alg:ALI-DPFL} based on Lemma \ref{lem:tau_0} and Lemma \ref{lem:tau_star}. Finally, in Section \ref{subsec:privacy_analysis}, we demonstrated that Algorithm \ref{alg:ALI-DPFL} satisfies the conditions for differential privacy.

\subsection{Motivation: Convergence Analysis of DPFL}
\label{subsec:ca}
To investigate the impact of $\tau$ on the convergence rate, we delve into the convergence of Algorithm \ref{alg:DPFL} and draw insights on selecting the optimal local iteration count $\tau^*$ based on the convergence upper bound. Prior to substantiating this, we establish several assumptions:

\assumption{$F_1, \cdots, F_N$ are all L-smooth\cite{watson1964smooth}: for all $\mathbf{v}$ and $\mathbf{w}, F_i(\mathbf{v}) \leq F_i(\mathbf{w})+(\mathbf{v}-\mathbf{w})^T \nabla F_i(\mathbf{w})+\frac{L}{2}\|\mathbf{v}-\mathbf{w}\|^2$.}\label{ass:l_smooth}

\assumption{$F_1, \cdots, F_N$ are all $\mu$-strongly convex\cite{polovinkin1996strongly}: for all $\mathbf{v}$ and $\mathbf{w}, F_i(\mathbf{v}) \geq F_i(\mathbf{w})+(\mathbf{v}-\mathbf{w})^T \nabla F_i(\mathbf{w})+\frac{\mu}{2}\|\mathbf{v}-\mathbf{w}\|^2$.}\label{ass:mu_strong_convex}

\assumption{The $L_2$ norm of stochastic gradients is uniformly bounded by the clipping bound \cite{MaxenceNoble2021DifferentiallyPF}, i.e.,$\left\|\nabla F_i\left(\mathbf{w}_t^i, s \right)\right\| \leq C$ for all $i=1, \cdots, N$ , $t=1, \cdots, T$  and $s \in \mathcal{B}^i$}\label{ass:grad_bound}

\theorem{Given the aforementioned assumptions, we derive the convergence upper bound for Algorithm \ref{alg:DPFL} as follows:}\label{theo:ca}

\begin{equation}\label{eq:ca}
    \mathbb{E}[F(\mathbf{w}_t)] - F^* \leq 
    h(\tau),
\end{equation}

where: 
\begin{align}\label{eq:h_tau}
    &h(\tau) \triangleq \frac{L(2+\eta\mu)\left(C^2+\frac{\sigma^2 C^2 d}{\hat{B}^2}\right)}{2T} \tau^2 \nonumber\\
    &\quad+ \frac{L\Delta_1 - 2L(2+\eta\mu)\left(C^2+\frac{\sigma^2C^2d}{\hat{B}^2}\right)}{2T}\tau + \nonumber\\
    & \frac{L \left( (2+\eta\mu)\left(C^2+\frac{\sigma^2 C^2 d}{\hat{B}^2}\right) + \frac{4}{\mu^2}+3C^2+\frac{2\Gamma}{\eta}+\frac{\sigma^2 C^2 d}{\hat{B}^2}\right)}{2T}.
\end{align}

Here $\frac{1}{\hat{B}} \triangleq \max_{i,t} \mathbb{E}\frac{1}{|\mathcal{B}^i_t|}$, $\Delta_1=\mathbb{E}\| \bar{\mathbf{w}}_1 - \mathbf{w}^* \|^2$. $\Gamma=F^* - \sum_{i=1}^{N} p_i F_i^*$, which is consistent with \cite{Li2019OnTheConvergence}, where a larger value of $\Gamma$ indicates that the data among different clients is more non-IID.

\textit{Proof Sketch:} We explore the relationship between $\mathbb{E}\| \bar{\mathbf{w}}_{t+1} - \mathbf{w}^{*} \|^2$ and $\mathbb{E}\| \bar{\mathbf{w}}_{t} - \mathbf{w}^{*} \|^2$, subsequently employing mathematical induction to establish an upper bound for $\mathbb{E}\| \mathbf{w}_{t} - \mathbf{w}^{*} \|^2$. Ultimately, leveraging Assumption \ref{ass:l_smooth}, we derive (\ref{eq:ca}). For a comprehensive proof, please refer to Appendix\protect\footnotemark \ref{append:proof_of_ca}.

\footnotetext{Due to the limited space, please access the appendix at arXiv:\\https://arxiv.org/abs/2308.10457}

\begin{lemma}\label{lem:tau_0}
Disregarding the limitation on the number of communication rounds, $\tau=1$ yields the fastest convergence.
\end{lemma}
\textit{Proof:} Observing Equation (\ref{eq:h_tau}), we note that $h(\tau)$ forms a quadratic function of $\tau$. The minimum value of $h(\tau)$ is attained when: 
\begin{equation*}
    \tau_0= 1 - \frac{\Delta_1}{2(2+\eta\mu)\left(C^2+\frac{\sigma^2 C^2 d}{\hat{B}^2}\right)}
\end{equation*}
As $0<\tau_0<1$, we conclude that $h(\tau)$ achieves its minimum value at $\tau=1$.

\begin{lemma}\label{lem:tau_star}
    Considering the constraint on the number of communication rounds, we can derive an optimal $\tau^*$:
\begin{align}\label{eq:optimal_tau_star}
    \tau^*=\sqrt{1 + \frac{\frac{4}{\mu^2} + 3C^2 +2 \Gamma T \mu + \frac{\sigma^2 C^2 d}{\hat{B}^2}}{\left(2 + \frac{1}{T}\right)\left(C^2 + \frac{\sigma^2 C^2 d}{\hat{B}^2}\right)}}
\end{align}
\end{lemma}
\textit{Proof:} The number of required communication rounds is determined by $\frac{T}{\tau}$. As this value is constant, we have:
\begin{align}
    h(\tau)=\frac{\tau}{T} \cdot G(\tau),
\end{align}
where:
\begin{align}\label{eq:g_tau}
    &G(\tau) \triangleq \frac{L(2+\eta\mu)\left(C^2+\frac{\sigma^2 C^2 d}{\hat{B}^2}\right)}{2} \tau \nonumber\\
    &\quad+ \frac{L\Delta_1 - 2L(2+\eta\mu)\left(C^2+\frac{\sigma^2C^2d}{\hat{B}^2}\right)}{2} + \nonumber\\
    & \frac{L \left( (2+\eta\mu)\left(C^2+\frac{\sigma^2 C^2 d}{\hat{B}^2}\right) + \frac{4}{\mu^2}+3C^2+\frac{2\Gamma}{\eta}+\frac{\sigma^2 C^2 d}{\hat{B}^2}\right)}{2\tau}.
\end{align}

When $\frac{T}{\tau}$ is a constant number, we minimize $G(\tau)$ to minimize $h(\tau)$. We observe that $G(\tau)$ is a parabolic function of $\tau$, and choosing excessively large or small values of $\tau$ would result in a large convergence bound, detrimental to the convergence of the algorithm. By minimizing $G(\tau)$, we obtain $\tau^*$ as given in Equation (\ref{eq:optimal_tau_star}).

\subsection{Overview}\label{subsec:overview}
\begin{figure}[htbp]
	\begin{center}
		\includegraphics[width=1.0\linewidth]{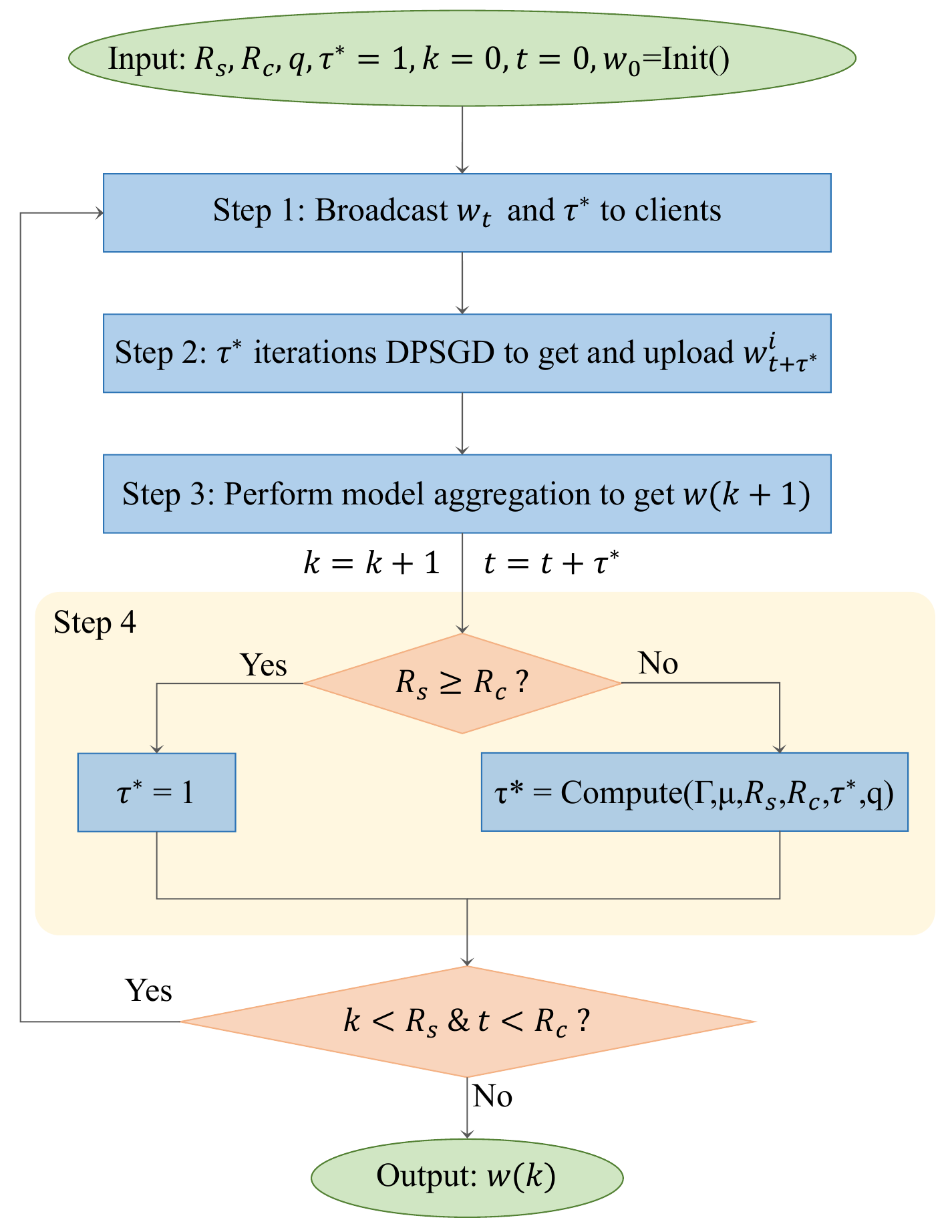}
		\caption{Workflow of ALI-DPFL}
		\label{fig:workflow}
	\end{center}
\end{figure}

As previously discussed, a fixed value of $\tau$ could impede the algorithm's rapid convergence. Hence, it is imperative to compute the optimal $\tau^*$ at each communication round. The process for computing $\tau^*$ is delineated in Figure \ref{fig:workflow}.

\begin{itemize}  
    \item \textbf{step 1}: The server disseminates the aggregated model $\mathbf{w}_t$ acquired from step 3 and the optimal local iterations $\tau^*$ obtained from step 4 to the clients.

    \item \textbf{step 2}: Each client concurrently executes $\tau^*$ iterations of DPSGD and uploads $\mathbf{w}^i_{t+\tau^*}$ to the central server.

    \item \textbf{step 3}: The server performs weighted aggregation on all models uploaded by the clients to derive the new server model for the subsequent round.
        
    \item \textbf{step 4}: The server computes $\tau^*$. If $R_s\geq R_c$, then $\tau^*=1$; otherwise, $\tau^*$ is computed using Equation (\ref{eq:optimal_tau_star}). The parameters required by Equation (\ref{eq:optimal_tau_star}) are calculated as follows:

    \begin{enumerate}
        \item Compute the strongly convex parameter as per Assumption \ref{ass:mu_strong_convex}:
        $$\mu \leftarrow \sum_{i=1}^N p_i \frac{ \|\nabla F_i(\mathbf{w}^i_t)-\nabla F_i(\mathbf{w}_t)\|}{\|\mathbf{w}^i_t-\mathbf{w}_t\|}$$
        
        \item Calculate $T$ based on $R_s, R_c$, and the previous round's $\tau^*$: 
        $$T \leftarrow \min \{ R_s\cdot \tau^*, R_c \} $$

        \item Determine $\hat{B}$ as per the definition: 
        $$\hat{B} \leftarrow \min_i q \cdot |\mathcal{D}_i|$$
    \end{enumerate}
\end{itemize}  

\subsection{Our Algorithm}
\label{subsec:algo_alidpfl}
\begin{algorithm}[ht]
\caption{Adaptive Local Iterations Differentially Private Federated Learning}\label{alg:ALI-DPFL}
\KwIn{loss function $F(\mathbf{w})$. Parameters: learning rate $\eta$, noise multiplier $\sigma$, privacy budget $\epsilon$, local iterations $\tau$, clipping bound for training $C$, sampling rate $q$, heterogeneous parameter $\Gamma$, maximum global rounds $R_s$, maximum total iterations $R_c$}
\KwOut{the final trained model $\mathbf{w}(k)$}
Initialize $k=0$, $t=0$, $\mathbf{w}_0=\text{Initial()}$\;
\While{$k<R_s$ and $t<R_c$}
{
\For{$i\in [1,2,\cdots,N]$ parallel}{
\For{$j \in [1,2,\cdots,\tau^*]$}
{
$t=t+1$\;
Sample randomly a batch $\mathcal{B}^i_t$ with sampling rate $q$\;
\For{$s\in \mathcal{B}^i_t$ }
{
$g^i_t(s) \leftarrow \nabla F_i(\mathbf{w}^i_t,s)$\; 
$\Bar{g}^i_t(s) \leftarrow g^i_t(s) / \max (1,\frac{\|g^i_t(s)\|}{C})$\; \label{alg2:line9}
}
$\widetilde{g}^i_t\leftarrow\sum\limits_{s \in {\mathcal{B}^i_t}}
\Bar{g}^i_t(s)+\mathcal{N}(0,\sigma^2 {C}^2 \mathbf{I}_d)$\;\label{alg2:line10}

$\mathbf{w}^i_{t+1}=\mathbf{w}^i_t-\eta \cdot \frac{\widetilde{g}^i_t}{|\mathcal{B}^i_t|}$\;
}
}
$\text{Receive } \mathbf{w}^i_{t+1}$\;
$k=k+1$\;
$\mathbf{w}(k)=\sum_{i=1}^N p_i \mathbf{w}^i_{t+1}$\;
\If{$R_s \geq R_c$}  
{  
    $\tau^*=1$\;  
}  
\Else  
{  
    $\tau^*$ = Compute($\Gamma,\mu,R_s,R_c,\tau^*,q$)  according to (\ref{eq:optimal_tau_star})\;  
}  
$\text{Broadcast } \mathbf{w}(k), \tau^* \text{ to each client}$ \;
}
\Return $\mathbf{w}(k)$
\end{algorithm}

\begin{algorithm}
\caption{Compute $R_c$}\label{algo:compute_Rc}
\KwIn{target privacy budget $\epsilon$, $\delta$, sampling rate $q$, noise multiplier $\sigma$}
Initialize $T_l=0$, $T_r=1e+10$, $\epsilon_m=0$\;
\While{$\epsilon_m - \epsilon < 0.01$}
{$T_m=\frac{T_l+T_r}{2}$\;
$\epsilon_m \gets \textbf{Compute privacy loss}(q,\sigma,T_m,\delta)$ by Theorem \ref{privacy theorem1} \; \label{line:alg3_line4}
\eIf{$\epsilon_m > \epsilon$}
{$T_r=T_m$}
{$T_l=T_m$}
}
\Return $T_l$
\end{algorithm}
Our objective is to adapt the value of $\tau$ to minimize either (\ref{eq:h_tau}) or (\ref{eq:g_tau}). This adjustment aims to bring $\mathbf{w}_t$ as close as feasible to $\mathbf{W}^*$, consequently facilitating swift convergence within resource constraints. As depicted in Algorithm \ref{alg:ALI-DPFL}, the disparity between ALI-DPFL and DPFL algorithms lies in the necessity to compute $\tau^*$ instead of relying on a fixed $\tau$.

\begin{itemize}  

    \item \textbf{lines 3-11 (Local DPSGD)}: During this phase, $N$ clients engage in parallel training. Initially, they receive $\mathbf{w}_t$ and $\tau^*$, then proceed with $\tau^*$ rounds of DPSGD iterations. In \textbf{line 6}, a sampling rate $q$ is utilized for Poisson sampling, drawing inspiration from \cite{IlyaMironov2017RnyiDP}, to obtain the data batch $\mathcal{B}^i_t$. Following this, in \textbf{lines 8-10}, the gradients of each sample within a single local iteration are clipped, then the gradients of the entire batch are summed, noise is added, and finally, the result is divided by the batch size $|\mathcal{B}^i_t|$. Subsequently, in \textbf{line 11}, gradient descent is performed locally using the processed gradients, and the updated model is uploaded to the server.
      
    \item \textbf{line 14 (Aggregation)}: Upon reception of models from each client, Algorithm \ref{alg:ALI-DPFL} proceeds to average the models based on weights.
      
    \item \textbf{lines 15-18 (Compute $\tau^*$)}: According to Lemma \ref{lem:tau_0} and Lemma \ref{lem:tau_star}, the approach to obtaining $\tau^*$ differs. When $R_s$ (server resources, i.e., maximum communication rounds) is greater than or equal to $R_c$ (client resources, i.e., maximum total local iterations), as per Lemma \ref{lem:tau_0}, it is established that $\tau^*=1$. Conversely, when $R_s$ is less than $R_c$, $\tau^*$ is computed based on Equation (\ref{eq:optimal_tau_star}).
    
\end{itemize}  

The value of $R_s$ can be pre-defined or controlled using methods such as time thresholds or communication bandwidth limits. Furthermore, the determination of $R_c$ can be achieved through Algorithm \ref{algo:compute_Rc} when establishing the parameters for differential privacy. Specifically, by defining the target privacy budget $\epsilon$, sampling rate $q$, and noise multiplier $\sigma$, the algorithm can obtain the corresponding total iteration rounds $R_c$ through a binary search.


\subsection{Privacy Analysis}\label{subsec:privacy_analysis}
Since each client's locally trained data samples share the same sampling rate, noise multiplier, and local iterations in each round, it follows that each client incurs an identical privacy loss in every round. Our analysis of each client's privacy loss is conducted from the viewpoint of an individual client. Subsequently, we investigate the privacy budget for the $i$-th client (the privacy loss analysis for the other clients follows the same procedure).
\begin{theorem}\label{privacy theorem1}(DP Privacy Loss of Algorithm \ref{alg:ALI-DPFL}). After $T$ local iterations, the $i$-th client DP Privacy budget of Algorithm \ref{alg:ALI-DPFL} satisfies:
\begin{equation}\label{eq:privacy_analysis}
\begin{split}
    (\epsilon^i_T,\delta)=(\sum_{t=0}^T \frac{1}{\alpha-1}\sum_{k=0}^{\alpha}\left(\begin{array}{l}
\alpha \\ i
\end{array}\right)(1-q)^{\alpha-k} q^{k}\\ \exp \left(\frac{k^{2}-k}{2 \sigma^{2}}\right)
+ \frac{\log 1/\delta}{\alpha-1},\delta)
\end{split}
\end{equation}
where $q$ is the sampling rate, $\sigma$ is noise multiplier and any integer $\alpha \geq 2$ .
\end{theorem}
$proof.$ Our privacy loss calculation is based on RDP \cite{IlyaMironov2017RnyiDP}. Initially, we utilize the sampling Gaussian theorem of RDP to compute the privacy cost of each round. Subsequently, we apply advanced combination techniques of RDP to aggregate the privacy cost across multiple rounds. Finally, we convert the resulting RDP privacy to DP.

\begin{definition}\label{privacy RDP privacy budget of SGM}
(RDP privacy budget of SGM\cite{IlyaMironov2019RnyiDP}). Let $SG_{q,\sigma}$, be the Sampled Gaussian Mechanism for some function $f$. If $f$ has sensitivity 1, $SG_{q,\sigma}$ satisfies $(\alpha,R)$-RDP whenever
\begin{equation}
R \leq \frac{1}{\alpha-1} \log \max(A_{\alpha}(q,\sigma),B_{\alpha}(q,\sigma))
\end{equation}
where:
\begin{equation}
\left\{\begin{array}{l}
A_{\alpha}(q, \sigma) \triangleq \mathbb{E}_{z \sim \vartheta_{0}}\left[\left(\vartheta(z) / \vartheta_{0}(z)\right)^{\alpha}\right] \\
B_{\alpha}(q, \sigma) \triangleq \mathbb{E}_{z \sim \vartheta}\left[\left(\vartheta_{0}(z) / \vartheta(z)\right)^{\alpha}\right]
\end{array}\right.
\end{equation}
with $\vartheta_{0} \triangleq \mathcal{N}\left(0, \sigma^{2}\right), \vartheta_{1} \triangleq \mathcal{N}\left(1, \sigma^{2}\right) \mbox { and } \vartheta \triangleq(1-q) \vartheta_{0}+q \vartheta_{1}$

Further, it holds $\forall(q,\sigma)\in(0,1], \mathbb{R}^{+*},A_{\alpha}(q,\sigma) \geq B_{\alpha}(q, \sigma) $. Thus, $ S G_{q, \sigma}  \text{ satisfies }  \left(\alpha, \frac{1}{\alpha-1} \log \left(A_{\alpha}(q, \sigma)\right)\right)$-RDP .

Finally, \cite{IlyaMironov2019RnyiDP} describes a procedure to compute $A_{\alpha}(q,\sigma)$ depending on integer $\alpha$.
\begin{equation}
A_{\alpha}=\sum_{k=0}^{\alpha}\left(\begin{array}{l}
\alpha \\ k
\end{array}\right)(1-q)^{\alpha-k} q^{k} \exp \left(\frac{k^{2}-k}{2 \sigma^{2}}\right)
\end{equation}
\end{definition}

\begin{definition} \label{privacy Composition of RDP}
(Composition of RDP\cite{IlyaMironov2017RnyiDP}). For two randomized mechanisms $f, g$ such that $f$ is $(\alpha,R_1)$-RDP and $g$ is $(\alpha,R_2)$-RDP the composition of $f$ and $g$ which is defined as $(X, Y )$(a sequence of results), where $ X \sim f $ and $Y \sim g$, satisfies $(\alpha,R_1+R_2)-RDP$
\end{definition}

From Definition~\ref{privacy RDP privacy budget of SGM} and  Definition~\ref{privacy Composition of RDP}, the following Lemma~\ref{privacy lamma1} is obtained.

\begin{lemma}\label{privacy lamma1} Given the sampling rate $q$ for each round of the local dataset and $\sigma$ as the noise factor, the total RDP privacy loss of the $i$-th client for local iterations $T$ for any integer $\alpha \geq 2$ is:
\begin{equation}
    \epsilon^{'}(\alpha)_T^i=\sum_{t=0}^T \frac{1}{\alpha-1}\sum_{k=0}^{\alpha}\left(\begin{array}{l}
\alpha \\ k
\end{array}\right)(1-q)^{\alpha-i} q^{k} \exp \left(\frac{k^{2}-k}{2 \sigma^{2}}\right)
\end{equation}
\end{lemma}

\begin{definition}\label{privacy Translation from RDP to DP}
(Translation from RDP to DP\cite{IlyaMironov2017RnyiDP}). if a randomized mechanism $f : D \rightarrow \mathbb{R}$  satisfies $(\alpha,\epsilon)$-RDP ,then it satisfies$(\epsilon +\frac{\log 1/\delta}{\alpha-1},\delta)$-DP where $0<\delta<1$.
\end{definition}

By Lemma~\ref{privacy lamma1} and Definition~\ref{privacy Translation from RDP to DP}, Theorem~\ref{privacy theorem1} is proved. 

\section{EXPERIMENT}\label{sec:experiment}
\addtolength{\topmargin}{0.03in}

We carried out a series of comparative experiments on the MNIST, FashionMnist, and Cifar10 datasets. The results of these experiments illustrate that our ALI-DPFL algorithm consistently delivers outstanding performance, enhancing accuracy across various ratios between $R_s$ and $R_c$.

\subsection{setup}
\subsubsection{baseline}
We compared the ALI-DPFL algorithm with three different baselines:

\begin{itemize}  
    \item \textbf{FedAvg\cite{mcmahan2017communication}+DP}: FedAvg, as the earliest Federated Learning algorithm, was executed with a fixed $\tau$. We replace SGD with DPSGD to ensure privacy protection while setting \textcolor{blue}{$\tau=\{1,2,3,5,10\}$} as the baselines.
    \item \textbf{PE-DPFL\cite{ShenX2022PEDPFL}}: This algorithm introduces a regularization term to the objective function and directly adds noise to the model before uploading. For our comparative experiments, we utilized hyperparameters epoch = 1 and $\tau=3$.
    \item \textbf{Adap DP-FL\cite{FuJ2022AdapDPFL}}: This algorithm dynamically adjusts the value of $\sigma$ based on whether the loss has decreased within four rounds. Similarly, for our comparative experiments, we employed hyperparameters epoch = 1 and $\tau=3$.
\end{itemize}
\subsubsection{dataset}
In our study, we conducted experiments on various $R_s$ and $R_c$ ratios under independent and identically distributed (IID) as well as non-independent and identically distributed (Non-IID) data distributions. We employed the following three real datasets: MNIST\cite{Yann1998MNIST}, FashionMNIST\cite{xiao2017fashionMNIST} and Cifar10\cite{AlexKrizhevsky2009LearningML}.




\begin{table*}[hbt]  
\centering  
\caption{(Small $R_s$) The average accuracy($\%$) on MNIST, Fashion MNIST and Cifar10.}  
\label{table:experiment_1}  
\begin{tabular}{|c|c c c|c c c|c c c|c c c|}  
\hline  
\multirow{2}{*}{} & \multicolumn{3}{c|}{$R_s=2R_c, \epsilon=1.55$} & \multicolumn{3}{c|}{$R_s=R_c, \epsilon=1.75$} & \multicolumn{3}{c|}{$R_s=\frac{1}{2}R_c, \epsilon=2$} & \multicolumn{3}{c|}{$R_s=\frac{1}{5}R_c, \epsilon=2.75$} \\  
\cline{2-13}  
 Datasets & MNIST & FMNIST & Cifar10 & MNIST & FMNIST & Cifar10 & MNIST & FMNIST & Cifar10 & MNIST & FMNIST & Cifar10 \\  
\hline  
PE-DPFL & 80.12 & 79.81 & 38.22 & 96.02 & 80.91 & 42.61 & 97.42 & 82.15 & 48.98 & 97.99 & 83.92 & 51.90  \\
Adap$\ $DP-FL & 85.83 & 80.11 & 39.12 & 94.21 & 81.31 & 46.12 & 95.41 & 81.30 & 47.39 & 96.98 & 82.05 & 49.96  \\
$\tau=1$\protect\footnotemark & 85.65 & 79.98 & 39.33 & 96.01 & 81.94 & 46.27 & 95.06 & 81.64 & 47.27 & 97.08 & 82.14 & 47.20  \\  
$\tau=2$ & 81.26 & 79.66 & 32.55 & 95.99 & 81.48 & 42.91 & 96.77 & 83.20 & 50.79 & 97.99 & 82.90 & 50.96  \\  
$\tau=3$ & 78.63 & 78.55 & 31.23 & 95.21 & 81.26 & 41.33 & 97.51 & 82.60 & 49.28 & 98.15 & 83.60 & 52.85  \\  
$\tau=5$ & 77.62 & 77.68 & 26.38 & 95.14 & 79.64 & 41.06 & 97.49 & 82.25 & 47.62 & 97.96 & 82.45 & 53.14  \\  
$\tau=10$ & 76.65 & 75.32 & 23.46 & 91.05 & 76.32 & 40.20 & 95.18 & 78.23 & 44.38 & 96.42 & 76.78 & 48.30  \\  
ALI-DPFL & $\textbf{86.10}$ & $\textbf{80.17}$ & $\textbf{39.82}$ 
& $\textbf{96.25}$ & $\textbf{82.02}$ & $\textbf{46.32}$ 
& $\textbf{97.58}$ & $\textbf{83.44}$ & $\textbf{51.92}$ 
& $\textbf{98.25}$ & $\textbf{84.07}$ & $\textbf{53.54}$ \\  
\hline  
\end{tabular}  
\end{table*}

\begin{table*}[hbt]  
\centering  
\caption{(Medium $R_s$) The average accuracy($\%$) on MNIST, Fashion MNIST and Cifar10.}  
\label{table:experiment_2}  
\begin{tabular}{|c|c c c|c c c|c c c|c c c|}  
\hline  
\multirow{2}{*}{} & \multicolumn{3}{c|}{$R_s=2R_c, \epsilon=1.75$} & \multicolumn{3}{c|}{$R_s=R_c, \epsilon=2$} & \multicolumn{3}{c|}{$R_s=\frac{1}{2}R_c, \epsilon=2.5$} & \multicolumn{3}{c|}{$R_s=\frac{1}{5}R_c, \epsilon=3.75$} \\  
\cline{2-13}  
 Datasets & MNIST & FMNIST & Cifar10 & MNIST & FMNIST & Cifar10 & MNIST & FMNIST & Cifar10 & MNIST & FMNIST & Cifar10 \\  
\hline  
PE-DPFL & 96.04 & 78.11 & 38.22 & 96.99 & 83.05 & 53.61 & 98.02 & 81.69 & 52.83 & 98.08 & 83.31 & 54.21  \\
Adap$\ $DP-FL & 93.08 & 80.21 & 43.53 & 97.12 & 82.70 & 55.21 & 97.02 & 82.88 & 52.77 & 97.42 & 82.05 & 52.32  \\
$\tau=1$ & 96.12 & 81.84 & 44.06 & 97.23 & 83.71 & 54.96 & 96.91 & 82.96 & 52.66 & 97.33 & 81.99 & 51.00  \\  
$\tau=2$ & 95.46 & 79.88 & 38.63 & 96.57 & 82.84 & 54.33 & \textbf{98.38} & 81.94 & 53.41 & 97.15 & 82.83 & 53.92  \\  
$\tau=3$ & 94.82 & 77.51 & 37.20 & 95.46 & 82.61 & 53.08 & 98.23 & 82.99 & 52.04 & 98.23 & 83.57 & 55.60  \\  
$\tau=5$ & 93.54 & 76.96 & 38.02 & 94.49 & 80.46 & 53.91 & 98.33 & 77.96 & 52.63 & 98.52 & 80.70 & 55.17  \\  
$\tau=10$ & 92.68 & 73.12 & 33.25 & 93.31 & 79.56 & 51.78 & 97.31 & 75.56 & 49.72 & 96.65 & 75.29 & 46.32  \\  
ALI-DPFL & $\textbf{96.14}$ & $\textbf{81.86}$ & $\textbf{44.12}$ 
& $\textbf{97.59}$ & $\textbf{83.88}$ & $\textbf{55.01}$ 
& 98.15 & $\textbf{83.58}$ & $\textbf{53.86}$ 
& $\textbf{98.82}$ & $\textbf{84.31}$ & $\textbf{55.67}$ \\  
\hline  
\end{tabular}  
\end{table*}  

\begin{table*}[hbt]  
\centering  
\caption{(Large $R_s$) The average accuracy($\%$) on MNIST, Fashion MNIST and Cifar10.}  
\label{table:experiment_3}  
\begin{tabular}{|c|c c c|c c c|c c c|c c c|}  
\hline  
\multirow{2}{*}{} & \multicolumn{3}{c|}{$R_s=2R_c, \epsilon=2$} & \multicolumn{3}{c|}{$R_s=R_c, \epsilon=2.5$} & \multicolumn{3}{c|}{$R_s=\frac{1}{2}R_c, \epsilon=3.25$} & \multicolumn{3}{c|}{$R_s=\frac{1}{5}R_c, \epsilon=5.25$} \\  
\cline{2-13}  
 Datasets & MNIST & FMNIST & Cifar10 & MNIST & FMNIST & Cifar10 & MNIST & FMNIST & Cifar10 & MNIST & FMNIST & Cifar10 \\  
\hline  
PE-DPFL & 92.94 & 81.81 & 42.22 & 97.43 & 81.90 & 50.31 & 98.12 & 81.84 & 52.91 & 98.12 & 83.94 & 56.70  \\
Adap$\ $DP-FL & 94.32 & 80.01 & 50.13 & 95.12 & 81.21 & 55.41 & 98.45 & 82.32 & 53.32 & 98.08 & 83.85 & 56.86  \\
$\tau=1$ & 94.92 & 82.53 & 49.99 & 97.57 & 84.02 & 56.41 & 97.93 & 83.39 & 54.33 & 96.98 & 83.70 & 53.37  \\  
$\tau=2$ & 94.46 & 81.49 & 44.54 & 97.46 & 82.92 & 50.11 & 98.58 & \textbf{83.46} & 54.96 & 97.89 & 83.72 & 55.56  \\  
$\tau=3$ & 92.98 & 81.56 & 41.14 & 97.32 & 82.73 & 50.59 & 98.61 & 83.24 & 53.08 & 98.11 & 83.75 & 56.84  \\  
$\tau=5$ & 92.86 & 78.48 & 41.25 & 97.49 & 82.50 & 50.47 & 98.85 & 81.67 & 53.91 & 98.33 & 81.72 & 55.21  \\  
$\tau=10$ & 91.98 & 75.32 & 38.85 & 96.58 & 81.72 & 47.98 & 97.21 & 79.65 & 51.78 & 97.42 & 81.25 & 54.55  \\  
ALI-DPFL & $\textbf{95.22}$ & $\textbf{82.58}$ & $\textbf{50.14}$ 
& $\textbf{97.74}$ & $\textbf{84.19}$ & $\textbf{56.76}$ 
& $\textbf{98.84}$ & 83.25 & $\textbf{55.17}$ 
& $\textbf{98.82}$ & $\textbf{84.58}$ & $\textbf{57.25}$ \\  
\hline  
\end{tabular}  
\end{table*}

\subsubsection{default parameters}
With a learning rate of $\eta=0.5$, we conducted experiments using ALI-DPFL with convolutional neural networks. We established the sampling rate $q$ at 0.015, with 10 clients participating in each training iteration. To simulate data heterogeneity, we employed a real heterogeneous setting \cite{TaoLin2020Ensemble} controlled by a Dirichlet distribution denoted as Dir($\beta$), where smaller $\beta$ values indicate higher data heterogeneity. We set the default value to $\beta=0.05$ \cite{WangJ2020FedNova}\cite{TaoLin2020Ensemble} and $\Gamma=10$. We conducted experiments using a neural network with 2 convolutional layers and 2 fully connected layers on three datasets.



\footnotetext{All fixed $\tau$ methods source from FedAvg+DP.}

\subsection{Performance}

In Lemma \ref{lem:tau_0}, we highlighted that $\tau=1$ represents the optimal choice for achieving the fastest convergence of the algorithm when the constraint of communication rounds is not taken into account. Previous studies analyzing convergence have primarily focused on the relationship between accuracy and the number of local iterations \cite{Li2019OnTheConvergence,HaoYu2019ParallelRestarted,KhaledA2020TighterTheory,SaiPK2020SCAFFLOD}. We have observed that the convergence bound consistently incorporates accuracy and the number of local iterations in the denominator. Holding accuracy constant, more iterations result in a smaller convergence bound. Selecting $\tau=1$ maximizes the number of iterations, aligning with our observation in Lemma \ref{lem:tau_0}. Based on this insight, we designed the following experiments:

$\textit{Condition I}$: When $R_s \geq R_c$, as per the aforementioned analysis, setting all values in the $\tau$ list to 1 yields the best model performance. Hence, we verified two cases: $R_s=R_c$ and $R_s=2R_c$, under three fixed conditions of $R_s$.

$\textit{Condition II}$: When $R_s<R_c$, setting all values in the $\tau$ list to 1 does not fully utilize local resources. Therefore, we examined two cases: $R_s=\frac{1}{2}R_c$ and $R_s=\frac{1}{5}R_c$, under three fixed conditions of $R_s$.

In our experiments, we compared the ALI-DPFL method with the fixed $\tau$ method, FedAvg\cite{mcmahan2017communication}+DP, PE-DPFL\cite{ShenX2022PEDPFL}, and Adap DP-FL\cite{FuJ2022AdapDPFL}. We conducted experiments for four different ratios: $R_s=R_c$, $R_s=2R_c$, $R_s=\frac{1}{2}R_c$, and $R_s=\frac{1}{5}R_c$, under three conditions: $R_s=\{158,231,317\}$. To accommodate the four cases of $R_s=\mathrm{n}\cdot R_c$ and the three values of $R_s=\{158,231,317\}$, we selected the following privacy budgets:$\epsilon=\{1.55,1.75,2,2.5,2.75,3.25,3.75,5.25\}$. These values support $\{79,177,317,598,775,1123,1585,2990\}$ iterations, approximately meeting the experimental requirements.

As depicted in Tables \ref{table:experiment_1}, \ref{table:experiment_2}, and \ref{table:experiment_3}, we evaluated the performance of these methods under Non-IID conditions (utilizing a Dirichlet distribution parameter of 0.05). Each entry in the tables represents the average result obtained after three replications. Our findings indicate that in the majority of cases, our method outperforms the other methods.

\begin{figure}[thb]  
    \centering  
    \begin{minipage}[b]{0.22\textwidth}  
        \centering  
        \includegraphics[width=\textwidth]{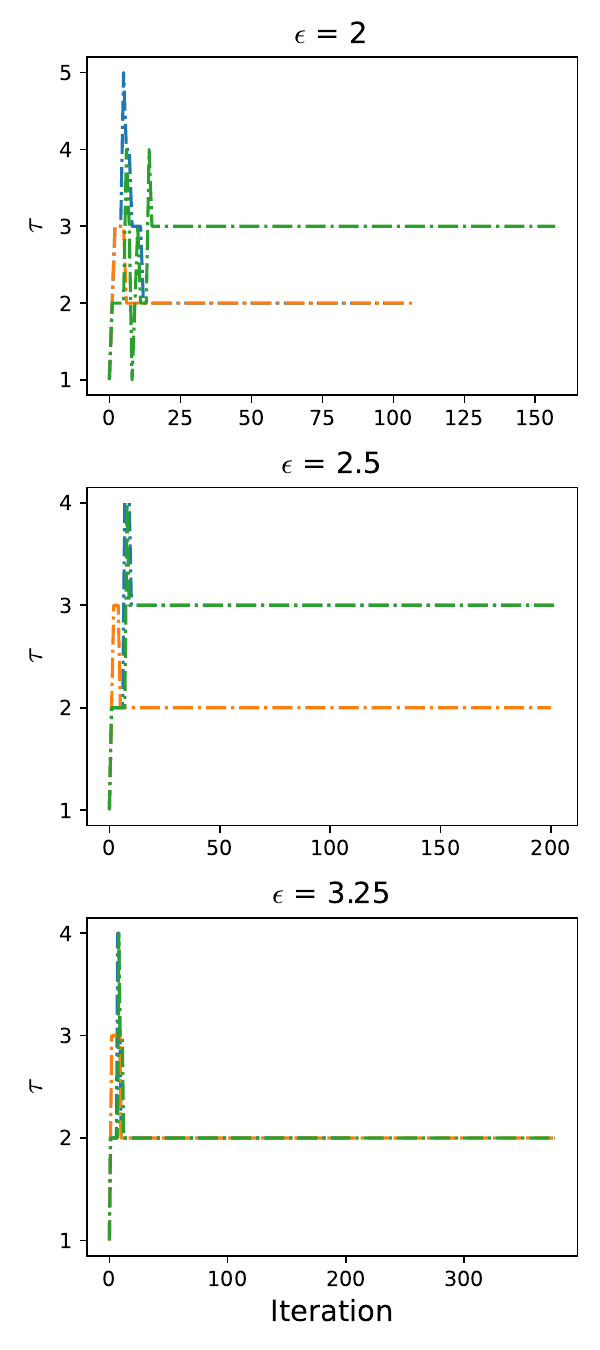}  
        \subcaption{$R_s=\frac{1}{2}R_c$}  
        \label{fig:rs12rc}  
    \end{minipage}  
    \hfill  
    \begin{minipage}[b]{0.22\textwidth}  
        \centering  
        \includegraphics[width=\textwidth]{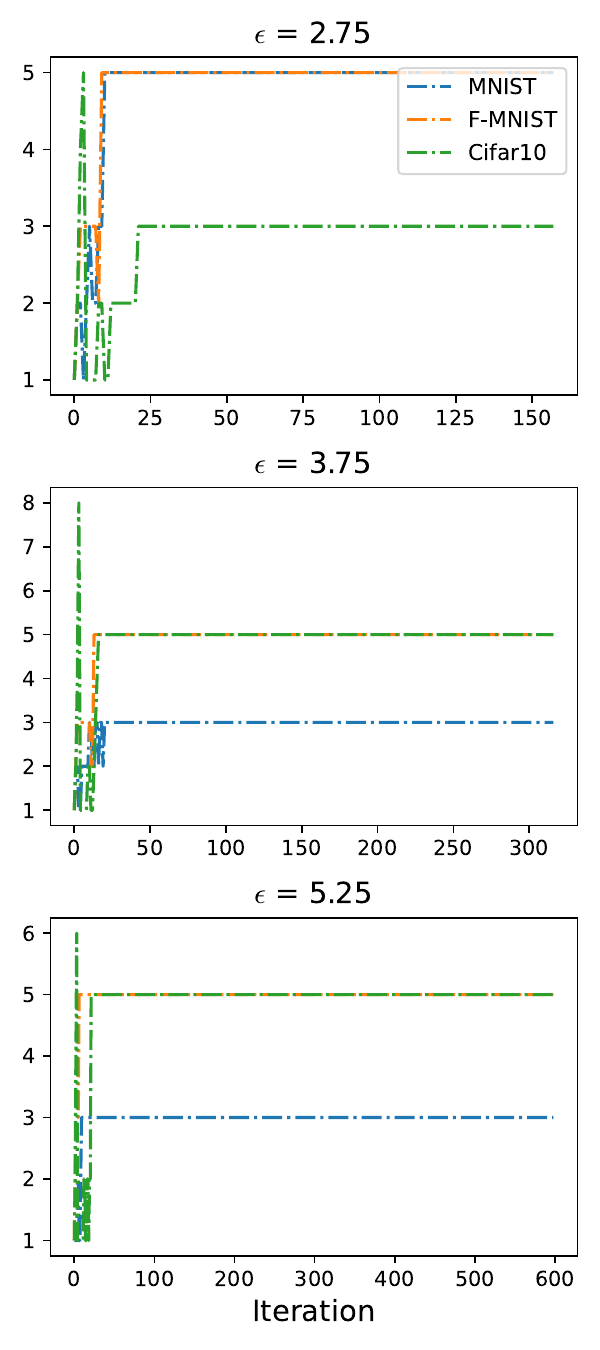}  
        \subcaption{$R_s=\frac{1}{5}R_c$}  
        \label{fig:rs15rc}  
    \end{minipage}  
  \caption{Adaptive $\tau$ value when $R_s < R_c$.} 
    \label{fig:adap_tau_list}  
\end{figure}

\begin{figure}[thb]
	\begin{center}
		\includegraphics[width=1.0\linewidth]{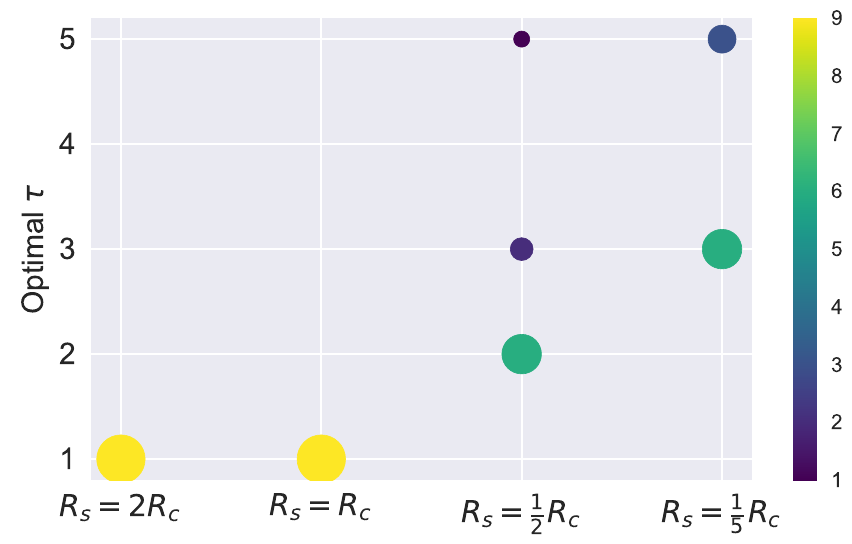}
		\caption{The frequencies of optimal fixed $\tau$ in Table \ref{table:experiment_1},\ref{table:experiment_2},\ref{table:experiment_3}}
		\label{fig:different_optimal_tau}
	\end{center}
\end{figure}

\subsection{Exploration the reason of outperform}
We counted the optimal $\tau$ values for each resource ratio in the fixed $\tau$ algorithm and illustrated them in Fig. \ref{fig:different_optimal_tau}. Our analysis revealed that when $R_s=\frac{1}{2}R_c$, the optimal $\tau$ values predominantly clustered around 2 and 3. Similarly, when $R_s=\frac{1}{5}R_c$, the optimal $\tau$ values were mainly concentrated in 3 and 5.

Subsequently, we depicted the $\tau$ lists for the ALI-DPFL algorithm where $R_s<R_c$ in Table \ref{table:experiment_1}, \ref{table:experiment_2}, and \ref{table:experiment_3} in Fig. \ref{fig:adap_tau_list}. Our observations indicated that initially, the value of $\tau^*$ exhibited fluctuations, but after several iterations, it stabilized to an appropriate $\tau$. Furthermore, when $R_s=\frac{1}{2}R_c$, the values in the $\tau$ list stabilized around 2 and 3, while for $R_s=\frac{1}{5}R_c$, the values in the $\tau$ list stabilized around 3 and 5. These findings suggest that ALI-DPFL effectively identifies the $\tau$ values conducive to rapid model convergence under varying resource ratios.

\subsection{Robustness with heterogeneous data}

\begin{table}[hbt]  
\centering  
\caption{The average accuracy($\%$) on heterogeneous setting at FasionMNIST dataset, $R_s=\frac{1}{3}R_c,\epsilon=2$.}  
\label{table:non_iid_experiments}  
\begin{tabular}{c|ccccc}  
\hline  
\textbf{} & \textbf{IID} & \textbf{Dir(0.05)} & \textbf{Dir(0.5)} & \textbf{Dir(1)} \\  
\hline  
PE-DPFL & 83.21 & 81.06 & 82.56 & 83.00 \\  
Adap DP-FL & 87.12 & 83.58 & 83.62 & 84.05 \\  
$\tau=1$ & 84.56 & 80.92 & 81.01 & 82.77 \\  
$\tau=2$ & 86.12 & 82.15 & 84.61 & 84.88 \\  
$\tau=3$ & 89.46 & 84.45 & 84.89 & 85.82 \\  
$\tau=5$ & 83.30 & 80.62 & 81.21 & 82.05 \\  
$\tau=10$ & 76.56 & 73.56 & 74.06 & 74.51 \\  
ALI-DPFL & \textbf{89.62} & \textbf{84.85} & \textbf{85.45} & \textbf{86.12} \\  
\hline  
\end{tabular}  

\end{table}  
To showcase the robustness of Algorithm \ref{alg:ALI-DPFL} in Non-IID data settings, we conducted experiments utilizing different data partitioning methods on the FashionMNIST dataset, with $R_s=\frac{1}{3}R_c$ and $\epsilon=2$. As depicted in Table \ref{table:non_iid_experiments}, the performance of the ALI-DPFL algorithm surpassed that of other methods in the cases of IID, Dir(0.05), Dir(0.5), and Dir(1). Correspondingly, we set the parameter $\Gamma$ to 0, 10, 5, and 1.

\section{Conclusion}\label{sec:conclusion}
\addtolength{\topmargin}{0.03in}
In this paper, we present an algorithm of Differential Private Federated Learning with Adaptive Local Iterations (ALI-DPFL), in the scenario where privacy budget and communication rounds are constrained. Through a theoretical convergence analysis of DPFL, we derive a convergence bound depending on the number of local iterations $\tau$, and improve the performance of federated learning with differential privacy by dynamically finding the optimal $\tau$. We formally prove the privacy of the proposed algorithm with the RDP technique, and conduct extensive experiments to demonstrate that ALI-DPFL significantly outperforms existent schemes in the resource-constraint scenario.

\section*{Acknowledgments}
This work is supported by the Natural Science Foundation of
Shanghai (Grant No. 22ZR1419100), the National Natural
Science Foundation of China Key Program (Grant No. 62132005), and  CAAI-Huawei MindSpore Open Fund (Grant No. CAAIXSJLJJ-2022-005A). This work is funded by the East China Normal University Graduate International Conference Special Fund.

\bibliographystyle{IEEEtran}
\bibliography{cite}
\clearpage

\appendix
\section{Appendix}
\subsection{Proof of Theorem \ref{theo:ca}}\label{append:proof_of_ca}
For the sake of convenience, we first establish some notations:

Let $\mathbf{w}_t^i$ represent the model of the $i$-th client at the $t$-th iteration. Let $\mathcal{I}_{\tau}$ denote the global aggregation step, which is defined as $\mathcal{I}_{\tau} = \{n \tau | n=1,2,\dots \}$. Based on Assumption \ref{ass:grad_bound}, the gradient descent equation can be expressed as:
\begin{align}\label{eq:gradient_decent_v_w}
\mathbf{v}_{t+1}^i = \mathbf{w}_t^i 
- \frac{\eta}{|\mathcal{B}^i_t|}
\left(
\sum_{s\in\mathcal{B}^i_t}
\nabla F_i(\mathbf{w}_t^i,s)
+ \sigma C \mathcal{N}(0,\mathbf{I}_d)
\right)
\end{align}
\begin{align}
\mathbf{w}_{t+1}^i= \begin{cases}\mathbf{v}_{t+1}^i & \text { if } t+1 \notin \mathcal{I}_\tau, \\ \sum_{i=1}^N p_i \mathbf{v}_{t+1}^i & \text { if } t+1 \in \mathcal{I}_\tau .\end{cases}
\end{align}
Here, we introduce $\mathbf{v}_{t+1}^i$ to represent the intermediate result of $\mathbf{w}_t^i$ after one iteration of gradient descent.

Inspired by \cite{Stich2019LocalSGDConvergesFastandCommunicatesLittle}, we introduce two virtual sequences in our analysis: $\bar{\mathbf{v}}_t = \sum_{i=1}^{N} p_i \mathbf{v}_{t}^i$ and $\bar{\mathbf{w}}_t = \sum_{i=1}^{N} p_i \mathbf{v}_{t}^i$. When $t+1 \notin \mathcal{I}_\tau$, $\bar{\mathbf{v}}_t = \bar{\mathbf{w}}_t$, but the server cannot access these two sequences. When $t+1 \in \mathcal{I}_\tau$, the server can access $\bar{\mathbf{w}}_t$.

Therefore, we have:

\begin{align}
\bar{\mathbf{v}}_{t+1} = \bar{\mathbf{w}}_t 
- \eta \sum_{i=1}^{N} 
\frac{1}{|\mathcal{B}^i_t|}
\left(
\sum_{s\in\mathcal{B}^i_t}
\nabla F_i(\mathbf{w}_t^i,s)
+ \sigma C \mathcal{N}(0,\mathbf{I}_d)
\right)
\end{align}

\subsubsection{Key lemma}

\lemma{(Results of one iteration)}\label{lem:result_one_iteration}. Assume Assumption \ref{ass:l_smooth}-\ref{ass:grad_bound}, we have:
\begin{equation}
    \mathbb{E}\| \bar{\mathbf{v}}_{t+1} - \mathbf{w}^* \|^2 \nonumber
    \leq  \left(1-\eta\mu \right) 
    \mathbb{E}\| \bar{\mathbf{w}}_{t} - \mathbf{w}^* \|^2
    + \eta^2 A
\end{equation}

since $\Delta_t= \mathbb{E}\|\bar{\mathbf{w}}_{t} - \mathbf{w}^*\|^2$, we have:
\begin{equation*}
    \Delta_{t+1} \leq \left(1-\eta\mu \right) \Delta_{t} + \eta^2 A
\end{equation*}
where $A = (\tau-1)^2(2+\eta\mu)\left( C^2 + \frac{\sigma^2 C^2 d}{\hat{B}}\right) +3C^2 +\frac{2 \Gamma}{\eta} + \frac{\sigma^2 C^2 d}{\hat{B}^2} $, $\Gamma= F^*- \sum_{i=1}^{N} p_iF_i^*$. 

We use mathematical induction to obtain $\Delta_t \leq \frac{v}{t}$ where $v=\max \{ \frac{\beta^2+A}{\beta\mu-1},\tau \Delta_1 \}$.

\textbf{STEP 1.} When $t=1$, the equation $\Delta_1 \leq v$ holds obviously.

\textbf{STEP 2.} We assume $\Delta_t \leq \frac{v}{t}$ holds.

\textbf{STEP 3.} on the condition that $\eta=\frac{\beta}{t}$, $\beta>\frac{1}{\mu}$, we know
\begin{align*}
    \Delta_{t+1} 
    &\leq \left(1- \eta\mu \right) \Delta_{t} + \eta^2 A \\
    &\leq  \left(1- \frac{\beta\mu}{t} \right) \frac{v}{t} + \frac{\beta^2+A}{t^2} \\
    &= \frac{(t-1)v}{t^2} + \left( \frac{\beta^2+A}{t^2} - \frac{\beta\mu-1}{t^2}v \right) \\
    &\leq \frac{t-1}{t^2}v \leq \frac{v}{t+1} \\
\end{align*}

Therefore, $\Delta_{t+1} \leq \frac{v}{t+1}$ holds, completing the proof by mathematical induction. Hence, $\Delta_t \leq \frac{v}{t}$ holds.

Then by the $L$-smoothness of $F(\cdot)$, and set $\beta=\frac{2}{\mu}$, $t \leftarrow T$

\begin{align}
    &\quad \mathbb{E}[F(\bar{\mathbf{w}}_T)]-F^* \leq \frac{L}{2}\Delta_T \nonumber\\
    &\leq \frac{L}{2T}\left( \frac{4}{\mu^2} + A + \tau\Delta_1 \right) \nonumber\\
    &\leq \frac{L}{2T}( \frac{4}{\mu^2} + \tau\Delta_1 + (\tau-1)^2(2+\eta\mu)\left( 
C^2+\frac{\sigma^2C^2d}{\hat{B}^2} \right) \nonumber\\
    &\quad +3C^2+\frac{2\Gamma}{\eta} + \frac{\sigma^2C^2d}{\hat{B}^2} ) \nonumber\\
    &= \frac{L(2+\eta\mu)\left( C^2 + \frac{\sigma^2C^2d}{\hat{B}^2} \right)}{2T}(\tau-1)^2
    \nonumber\\
    &\quad +\frac{L}{2T}\left( \frac{4}{\mu^2} + \tau\Delta_1 +3C^2 +\frac{2\Gamma}{\eta} + \frac{\sigma^2C^2d}{\hat{B}^2} \right)\nonumber\\
    &= h(\tau)
\end{align}

where
\begin{align}
    h(\tau) &\triangleq \frac{L(2+\eta\mu)\left(C^2+\frac{\sigma^2 C^2 d}{\hat{B}^2}\right)}{2T} \tau^2 \nonumber\\
    &\quad + \frac{L\Delta_1 - 2L(2+\eta\mu)\left(C^2+\frac{\sigma^2C^2d}{\hat{B}^2}\right)}{2T}\tau + \nonumber\\
    & \frac{L \left( (2+\eta\mu)\left(C^2+\frac{\sigma^2 C^2 d}{\hat{B}^2}\right) + \frac{4}{\mu^2}+3C^2+\frac{2\Gamma}{\eta}+\frac{\sigma^2 C^2 d}{\hat{B}^2}\right)}{2T}
\end{align}

\subsubsection{Proof of Lemma \ref{lem:result_one_iteration}}\hfill

According to (\ref{eq:gradient_decent_v_w}), 
\begin{align}\label{eq:start_of_CA}
&\quad \left\| \bar{\mathbf{v}}_{t+1}-\mathbf{w}^{*} \right\|^2 \nonumber\\
&= \| \bar{\mathbf{w}}_{t} -\mathbf{w}^{*} \nonumber\\
&\quad -\eta \sum_{i=1}^N p_i \frac{1}{|\mathcal{B}^i_t|}
\left(
\sum_{s\in\mathcal{B}^i_t}
\nabla F_i(\mathbf{w}_t^i,s)
+ \sigma C \mathcal{N}(0,\mathbf{I}_d)
\right)
\|^2\nonumber\\
&= \left\| \bar{\mathbf{w}}_{t}-\mathbf{w}^{*} \right\|^2 \nonumber\\
&+2\eta \underbrace{\langle\mathbf{w}^{*}-\bar{\mathbf{w}}_{t},
\sum_{i=1}^N p_i \frac{1}{|\mathcal{B}^i_t|}
\left(
\sum_{s\in\mathcal{B}^i_t}
\nabla F_i(\mathbf{w}_t^i,s)
+ \sigma C \mathcal{N}(0,\mathbf{I}_d)
\right)\rangle}_{\mathcal{A}_1} \nonumber\\
&+ \eta^2 \underbrace{\| \sum_{i=1}^N p_i \frac{1}{|\mathcal{B}^i_t|}
\left(
\sum_{s\in\mathcal{B}^i_t}
\nabla F_i(\mathbf{w}_t^i,s)
+ \sigma C \mathcal{N}(0,\mathbf{I}_d)
\right) \|^2}_{\mathcal{A}_2}
\end{align}

By the convexity of $\|\cdot\|^2$,

\begin{align*}
    \mathcal{A}_2 &= \left\| \sum_{i=1}^N p_i \frac{1}{|\mathcal{B}^i_t|}
\left(
\sum_{s\in\mathcal{B}^i_t}
\nabla F_i(\mathbf{w}_t^i,s)
+ \sigma C \mathcal{N}(0,\mathbf{I}_d)
\right) \right\|^2 \\
    &\leq \sum_{i=1}^N p_i \left\|  \frac{1}{|\mathcal{B}^i_t|}
\left(
\sum_{s\in\mathcal{B}^i_t}
\nabla F_i(\mathbf{w}_t^i,s)
+ \sigma C \mathcal{N}(0,\mathbf{I}_d)
\right) \right\|^2 \\ \\
    &\leq \sum_{i=1}^N p_i \left\|\frac{1}{|\mathcal{B}^i_t|} \sum_{s\in\mathcal{B}^i_t} \nabla F_i(\mathbf{w}_t^i,s) \right\|^2 \\
    &\quad+  \sum_{i=1}^N p_i \left\| \frac{1}{|\mathcal{B}^i_t|} \sigma C \mathcal{N}(0,\mathbf{I}_d) \right\|^2 \\
    & (\text{From the Cauchy Schwartz inequality.})
\end{align*}
Take the expectation of $\mathcal{A}_2$:
\begin{align*}
    \mathbb{E}\mathcal{A}_2 
    \leq C^2 + \frac{\sigma^2 C^2 d}{\hat{B}^2}
\end{align*}

where $\frac{1}{\hat{B}} = \max_{i,t} \mathbb{E} \frac{1}{|\mathcal{B}^i_t|}$. 

\begin{align*}
    \mathcal{A}_1 &=
    -\langle\bar{\mathbf{w}}_{t}-\mathbf{w}^{*},
\sum_{i=1}^N p_i \frac{1}{|\mathcal{B}^i_t|}
\left(
\sum_{s\in\mathcal{B}^i_t}
\nabla F_i(\mathbf{w}_t^i,s)
+ \sigma C \mathcal{N}(0,\mathbf{I}_d)
\right)\rangle \\
    &= \underbrace{-\langle\bar{\mathbf{w}}_{t}-\mathbf{w}^{*},
\sum_{i=1}^N p_i \frac{1}{|\mathcal{B}^i_t|}
\sum_{s\in\mathcal{B}^i_t}
\nabla F_i(\mathbf{w}_t^i,s)
\rangle}_{\mathcal{B}_1} \\
&\quad \underbrace{-\langle\bar{\mathbf{w}}_{t}-\mathbf{w}^{*},
\sum_{i=1}^N p_i \frac{1}{|\mathcal{B}^i_t|}
\sigma C \mathcal{N}(0,\mathbf{I}_d)
\rangle}_{\mathcal{B}_2} \\
\end{align*}

Because the noise has a mean of 0:
\begin{equation}
    \mathbb{E}\mathcal{B}_2 = 0
\end{equation}
Add a zero term to $\mathcal{B}_1$:
\begin{align*}
    \mathcal{B}_1 &= -\sum_{i=1}^N p_i \langle\bar{\mathbf{w}}_{t} -\mathbf{w}_t^i + \mathbf{w}_t^i -\mathbf{w}^{*},
\frac{1}{|\mathcal{B}^i_t|}
\sum_{s\in\mathcal{B}^i_t}
\nabla F_i(\mathbf{w}_t^i,s)\rangle \\
&= \underbrace{- \sum_{i=1}^N p_i \langle\bar{\mathbf{w}}_{t} -\mathbf{w}_t^i,
\frac{1}{|\mathcal{B}^i_t|}
\sum_{s\in\mathcal{B}^i_t}
\nabla F_i(\mathbf{w}_t^i,s)\rangle}_{\mathcal{C}_1} \\
&\quad \underbrace{- \sum_{i=1}^N p_i \langle\mathbf{w}_t^i -\mathbf{w}^{*},
\frac{1}{|\mathcal{B}^i_t|}
\sum_{s\in\mathcal{B}^i_t}
\nabla F_i(\mathbf{w}_t^i,s)\rangle}_{\mathcal{C}_2} \\
\end{align*}

By Cauchy-Schwarz inequality and AM-GM inequality, we have
\begin{align*}
    \mathcal{C}_1 &\leq \frac{1}{\eta} \sum_{i=1}^N p_i \| \bar{\mathbf{w}}_{t} -\mathbf{w}_t^i \|^2
    + \eta \sum_{i=1}^N p_i \left\|  \frac{1}{|\mathcal{B}^i_t|}
\sum_{s\in\mathcal{B}^i_t}
\nabla F_i(\mathbf{w}_t^i,s) \right\|^2\\
    &\leq \frac{1}{\eta} \sum_{i=1}^N p_i \| \bar{\mathbf{w}}_{t} -\mathbf{w}_t^i \|^2 +\eta C^2
\end{align*}
According to Assumption \ref{ass:mu_strong_convex}, we get
\begin{align*}
    \mathcal{C}_2 &\leq -\sum_{i=1}^N p_i \langle\mathbf{w}_t^i -\mathbf{w}^{*},
\frac{1}{|\mathcal{B}^i_t|}
\sum_{s\in\mathcal{B}^i_t}
\nabla F_i(\mathbf{w}_t^i,s)\rangle \\
&\leq -\sum_{i=1}^{N} p_i \left( F_i(\mathbf{w}_t^i) - F_i(\mathbf{w}^*) \right)
- \sum_{i=1}^N p_i \frac{\mu}{2}\| \mathbf{w}_t^i - \mathbf{w}^* \|^2\\
&= -\sum_{i=1}^{N} p_i \left( F_i(\mathbf{w}_t^i) - F(\mathbf{w}^*) + F(\mathbf{w}^*) - F_i(\mathbf{w}^*) \right)\\
&\quad -\frac{\mu}{2} \sum_{i=1}^N p_i \| \mathbf{w}_t^i - \mathbf{w}^* \|^2 \\
&= -\sum_{i=1}^{N} p_i \left( F_i(\mathbf{w}_t^i) - F(\mathbf{w}^*)\right) + \sum_{i=1}^{N} p_i \left(F_i(\mathbf{w}^*) -F(\mathbf{w}^*) \right)\\
&\quad-\frac{\mu}{2} \sum_{i=1}^N p_i \| \mathbf{w}_t^i - \mathbf{w}^* \|^2 \\
&\leq -\underbrace{\sum_{i=1}^{N} p_i \left( F_i(\mathbf{w}_t^i) - F(\mathbf{w}^*) \right)}_{\mathcal{D}_1} - \frac{\mu}{2} \sum_{i=1}^N p_i \| \mathbf{w}_t^i - \mathbf{w}^* \|^2 + \Gamma
\end{align*}
where $\Gamma \triangleq \sum_{i=1}^N p_i \left( F(\mathbf{w}^*) - F_i(\mathbf{w}^*) \right)$.

Process the $\mathcal{D}_1$
\begin{align*}
    \mathcal{D}_1 &= \sum_{i=1}^{N} p_i \left( F_i(\mathbf{w}_t^i) - F(\mathbf{w}^*) \right) \\
    & = \sum_{i=1}^{N} p_i \left( F_i(\mathbf{w}_t^i) - F_i(\bar{\mathbf{w}}_t) \right)
    + \sum_{i=1}^{N} p_i \left( F_i(\bar{\mathbf{w}}_t) - F(\mathbf{w}^*) \right)\\
    &\geq \sum_{i=1}^{N} p_i \langle \nabla F_i(\bar{\mathbf{w}_t}) , \mathbf{w}_t^i - \bar{\mathbf{w}}_t \rangle + \left( F(\bar{\mathbf{w}}_t) - F(\mathbf{w}^*) \right)\\
    &\text{(from the Assumption \ref{ass:mu_strong_convex})}\\
    &\geq -\frac{1}{2} \sum_{i=1}^{N} p_i \left[ \eta \| F_i(\bar{\mathbf{w}}_t)  \|^2 + \frac{1}{\eta}
    \| \mathbf{w}_t^i - \bar{\mathbf{w}}_t \|^2
    \right] \nonumber\\
    &\quad + \left( F(\bar{\mathbf{w}}_t) - F(\mathbf{w}^*) \right)\\
    &\text{(from the AM-GM inequality)}\\
    &\geq - \sum_{i=1}^{N} p_i \left[ \eta L \left( F_i(\bar{\mathbf{w}}_t) - F(\mathbf{w}^*) \right) + \frac{1}{2\eta} \| \mathbf{w}_t^i - \bar{\mathbf{w}}_t \|^2 \right] \\
    &+ \left( F(\bar{\mathbf{w}}_t) - F(\mathbf{w}^*) \right)\\
    &\text{(from the Assumption \ref{ass:l_smooth})}\\
    &= (1-\eta L)\left( F(\bar{\mathbf{w}}_t) - F(\mathbf{w}^*) \right) 
    + \frac{1}{2\eta} \sum_{i=1}^N p_i \| \mathbf{w}_t^i - \bar{\mathbf{w}}_t \|^2\\
\end{align*}

Hence, we get
\begin{align*}
    \mathcal{B}_1 &\leq  \frac{1}{\eta} \sum_{i=1}^N p_i \| \bar{\mathbf{w}}_{t} -\mathbf{w}_t^i \|^2 +\eta C^2  \\
    & + (\eta L - 1)\left( F(\bar{\mathbf{w}}_t) - F(\mathbf{w}^*) \right) 
    - \frac{1}{2\eta} \| \mathbf{w}_t^i - \bar{\mathbf{w}}_t \|^2 \\
    & - \frac{\mu}{2} \sum_{i=1}^N p_i \| \mathbf{w}_t^i - \mathbf{w}^* \|^2 + \Gamma\\
    &\leq -\frac{\mu}{2} \| \bar{\mathbf{w}}_t - \mathbf{w}^* \|^2
    + (\frac{1}{\eta}+\frac{\mu}{2}) \sum_{i=1}^N p_i \| \bar{\mathbf{w}}_t - \mathbf{w}_t^i \|^2 + \eta C^2 + \Gamma
\end{align*}

Comprehensive equation $\mathcal{A}-\mathcal{D}$, take exception to (\ref{eq:start_of_CA}):
\begin{align}\label{eq:middle_result_of_CA}
    &\quad \mathbb{E}\|\bar{\mathbf{v}}_{t+1} - \mathbf{w}^* \|^2 
    =  \mathbb{E}\|\bar{\mathbf{w}}_{t} - \mathbf{w}^* \|^2 
    + 2\eta\mathbb{E}\mathcal{A}_1 + \eta^2\mathbb{E}\mathcal{A}_2\nonumber\\
    &\leq \mathbb{E}\|\bar{\mathbf{w}}_{t} - \mathbf{w}^* \|^2 
    + \eta^2 \left( C^2 + \frac{\sigma^2 C^2 d}{\hat{B}^2} \right) \nonumber\\
    &\quad+ 2\eta ( -\frac{\mu}{2} \mathbb{E}\| \bar{\mathbf{w}}_{t} - \mathbf{w}^* \|^2 \nonumber\\
    &\quad + \left( \frac{1}{\eta} + \frac{\mu}{2} \right) \sum_{i=1}^N p_i \mathbb{E}\| \bar{\mathbf{w}}_{t} - \mathbf{w}_t^i \|^2 +\eta C^2 +\Gamma ) \nonumber\\
    &= (1-\eta\mu)\mathbb{E}\|\bar{\mathbf{w}}_{t} - \mathbf{w}^* \|^2 
    + (2+\eta\mu) \sum_{i=1}^N p_i \mathbb{E}\|\bar{\mathbf{w}}_{t} - \mathbf{w}_t^i \|^2 \nonumber\\
    &\quad +3\eta^2 C^2 + 2 \eta \Gamma + \frac{\eta^2 \sigma^2 C^2 d}{\hat{B}^2}
\end{align}

Then we do such process:
\begin{align}\label{eq:ordinary_lemma3}
    &\quad\sum_{i=1}^N p_i \mathbb{E}\|\bar{\mathbf{w}}_{t} - \mathbf{w}_t^i \|^2\nonumber\\ 
    &= \sum_{i=1}^N p_i \mathbb{E}\|(\mathbf{w}_{t}^{i} - \bar{\mathbf{w}}_{t_0})
    - (\bar{\mathbf{w}}_{t} - \bar{\mathbf{w}}_{t_0}) \|^2 \nonumber\\
    &\leq \sum_{i=1}^N p_i \mathbb{E} \sum_{t=t_0}^{t-1} (\tau-1) \eta^2
    \left\| \frac{1}{|\mathcal{B}^i_t|}
\left(
\sum_{s\in\mathcal{B}^i_t}
\nabla F_i(\mathbf{w}_t^i,s)
+ \sigma C \mathcal{N}(0,\mathbf{I}_d)
\right) \right\|^2 \nonumber\\
&\leq \sum_{i=1}^N p_i \eta^2 (\tau-1)^2 \left( \mathbb{E} 
    \left\| \frac{1}{|\mathcal{B}^i_t|}
\sum_{s\in\mathcal{B}^i_t}
\nabla F_i(\mathbf{w}_t^i,s) 
+ \right\|^2 \frac{\sigma^2 C^2 d}{\hat{B}^2} \right) \nonumber\\
&\leq \eta^2 (\tau-1)^2 \left( C^2 + \frac{\sigma^2 C^2 d}{\hat{B}^2} \right)
\end{align}

Combine Equation (\ref{eq:start_of_CA}) and Equation (\ref{eq:ordinary_lemma3}):
\begin{equation}
    \mathbb{E}\|\bar{\mathbf{v}}_{t+1} - \mathbf{w}^* \|^2 
    \leq (1-\eta\mu) \mathbb{E}\|\bar{\mathbf{w}}_{t} - \mathbf{w}^* \|^2 + \eta A
\end{equation}
where
\begin{equation}
    A=(\tau-1)^2(2+\eta\mu)\left( C^2 + \frac{\sigma^2 C^2 d}{\hat{B}^2} +3C^2 + \frac{2\Gamma}{\eta} +\frac{\sigma^2 C^2 d}{\hat{B}^2} \right)
\end{equation}

\end{document}